\documentclass[10pt, conference, compsocconf]{IEEEtran}
\usepackage{tablefootnote}
\ifCLASSINFOpdf
\usepackage[pdftex]{graphicx}
\else
\fi

\usepackage[cmex10]{amsmath}

\usepackage[tight,footnotesize]{subfigure}

\begin{document}
\title{Very Long Term Field of View Prediction for 360-degree Video Streaming}
\author{\IEEEauthorblockN{Chenge Li, Weixi Zhang, Yong Liu, Yao Wang}
\IEEEauthorblockA{Department of Electrical and Computer Engineering, Tandon School of Engineering\\
New York University, Brooklyn, NY 11201, USA\\
\{cl2840, wz1219, yongliu, yw523\}@nyu.edu}

}

\maketitle

\begin{abstract}
360-degree videos have gained increasing popularity in recent years with the developments and advances in Virtual Reality (VR) and Augmented Reality (AR) technologies. In such applications, a user only watches a video scene within a field of view (FoV) centered in a certain direction. Predicting the future FoV in a long time horizon (more than seconds ahead)  can help save bandwidth resources in on-demand video streaming while minimizing video freezing in networks with significant bandwidth variations. In this work, we treat the FoV prediction as a sequence learning problem, and propose to predict the target user's future FoV not only based on the user's own past FoV center trajectory but also other users' future FoV locations. We propose multiple prediction models based on two different FoV representations: one using FoV center trajectories and another using equirectangular heatmaps that represent the FoV center distributions. Extensive evaluations with two public datasets demonstrate that the proposed models can significantly outperform benchmark models, and other users' FoVs are very helpful for improving long-term predictions. 
\end{abstract}


%\IEEEpeerreviewmaketitle

\section{Introduction}
Many VR/AR applications involve streaming of 360-degree videos, ranging from pre-recorded 360-degree scenes to live events captured from a camera array to physical \& virtual interactive environments. For on-demand streaming of precoded video content, in order to absorb significant bandwidth fluctuation between a server and a client, the client typically prefetches future video segments and store them in a display buffer. The prefetching buffer is typically more than 5 seconds long, in order to prevent ``video freezing" when the bandwidth suddenly drops. For 360$^\circ$ video streaming, this means that the client or the server has to predict the viewer's FoV far into the future, so that appropriate portions of the future video segments can be delivered.  The further into the future an accurate prediction can be performed, the more robust will be the streaming system to the bandwidth fluctuations. 

We treat the FoV prediction problem as a sequence prediction problem and propose two groups of prediction models: trajectory-based approach and heatmap-based approach. In the first group, we predict the mean and standard deviation (STD) of the FoV centers in future seconds. This approach is developed for \textit{viewport-based streaming} systems, where the client can request a single viewport for a future second based on the predicted FoV mean and standard deviation. We propose a Long Short-Term Memory (LSTM) sequence to sequence model and compare it with several benchmark models. Furthermore, assuming the video server has stored the FoV trajectories of other viewers who have watched the same video, we consider multiple strategies for using other viewers' trajectories to help the prediction of a target user's future trajectory. 

In heatmap-based approach, where we represent the FoV distribution for all frames within a second as a heatmap and predict the heatmaps in future seconds. Such an approach is intended for \textit{tile-based streaming} systems, where the client can request multiple tiles for a future second based on the predicted heatmap. We propose a convolutional LSTM\cite{xingjian2015convolutional} based model to predict the heatmap sequence of the target viewer in the future from the viewer's heatmap sequence in the past. We further consider using the heatmap sequences of other users and saliency maps derived from the video sequence to help predict the target user's future heatmaps.

\section{Related Work}
\label{relatedworks}
The FoV prediction algorithms in the literature can be categorized into two classes: trajectory based and content based. \cite{bao2016shooting} proposed to use linear regression and a 3-layer MLP to predict the future FoV center locations. Compared with our setting, their prediction horizon is very short: only $100 \sim 500$ ms. \cite{fan2017fixation} proposed a fixation prediction network that concurrently leverages past FoV locations and video content features to predict the FoV trajectory or tile-based viewing probability maps in the next $n$ frames. In \cite{shanghaitech}, an LSTM is used to encode the history of the FoV scan path and the hidden state features are combined with the visual features to do prediction up to 1 second ahead. A more recent work \cite{xu2018predicting} proposed two deep reinforcement learning models: one offline model is first used to estimate the heatmap of potential FoV for each frame based on the visual features only, an online model is then used to predict the head movement based on the past observed head locations as well as the heatmaps from the offline model.

Several prior studies also exploited the \textbf{cross-users behaviors} instead of only the target user's historical trajectories. \cite{ban2018cub360} and \cite{tilebased} combined a linear regression (LR) model with KNN clustering. From historical trajectories of head movements, FoV center is firstly predicted using a linear regression model, then the K nearest fixations of other users around the LR result are found to improve the prediction accuracy. A very recent work \cite{xie2018cls} first used the density-based clustering algorithm DBSCAN to group users in the server, then on the client end, an SVM classifier was used to predict the user's class to obtain the corresponding viewing probabilities from the clusters.

A key difference of our work from the prior related studies is that we focus on predicting the FoV for a much longer time horizon (1 to 10 seconds vs $\leq$ 1 second), which enables the streaming system to prefetch future video segments multiple seconds ahead, and to be more robust to the bandwidth fluctuations. 

%------------------------------------------------------------------------- 
\section{Trajectory-Based Prediction}
\subsection{Prediction based on user's own past}
\label{seq2seq section}
In related works such as \cite{bao2016shooting}  \cite{shanghaitech}, future FoVs are predicted by unrolling a single trained LSTM model. However, such single LSTM model is appropriate only if the input data types and data distribution in the past are the same as that in the future. To accommodate different input data types and distributions, we adopt the neural machine translation architecture \cite{sutskever2014sequence}, the seq2seq model, (see lower branch of Figure \ref{mlpmixing_fig}). We use an LSTM to encode the past trajectory $x_t$ in time $t=1,2,..., T$, and use the last hidden state $h_T$ and memory state $c_T$ as the representation of the past. We then use another LSTM initialized by $h_T$ and $c_T $ and an initial input $(\mu_T, \sigma_T)$ (mean and STD of $x_T$), to generate the hidden and memory states in future times $t=T+1,T+2, ..., T+L$. The final prediction $y_t$ (i.e. $(\mu_t, \sigma_t)$) at time $t$ is obtained using one projection layer from the hidden state $h_t$. The LSTM encoder uses the frame-level trajectory in each past second as the input (i.e., the sequence of $(x,y,z)$ coordinates of the FoV centers in all the frames in a second). The LSTM decoder uses the predicted mean $\mu_{t-1}$ and STD $\sigma_{t-1}$ for time $t-1$ as the input for time $t$. The encoder and decoder are trained together to minimize the prediction error for the $\mu$ and $\sigma$ for future $L$ seconds.

\subsection{Prediction with the help of others' FoVs}
In video-on-demand applications, the same video is often watched by many users. For a new streaming session, we can predict the target user's future trajectory based on this user's past trajectory as well as the trajectories of other users who have watched the same video before. We treat other users as experts, and learn how to efficiently utilize their ``guidance". 

\begin{figure}[!t]
\center
   \includegraphics[width=1.0\linewidth]{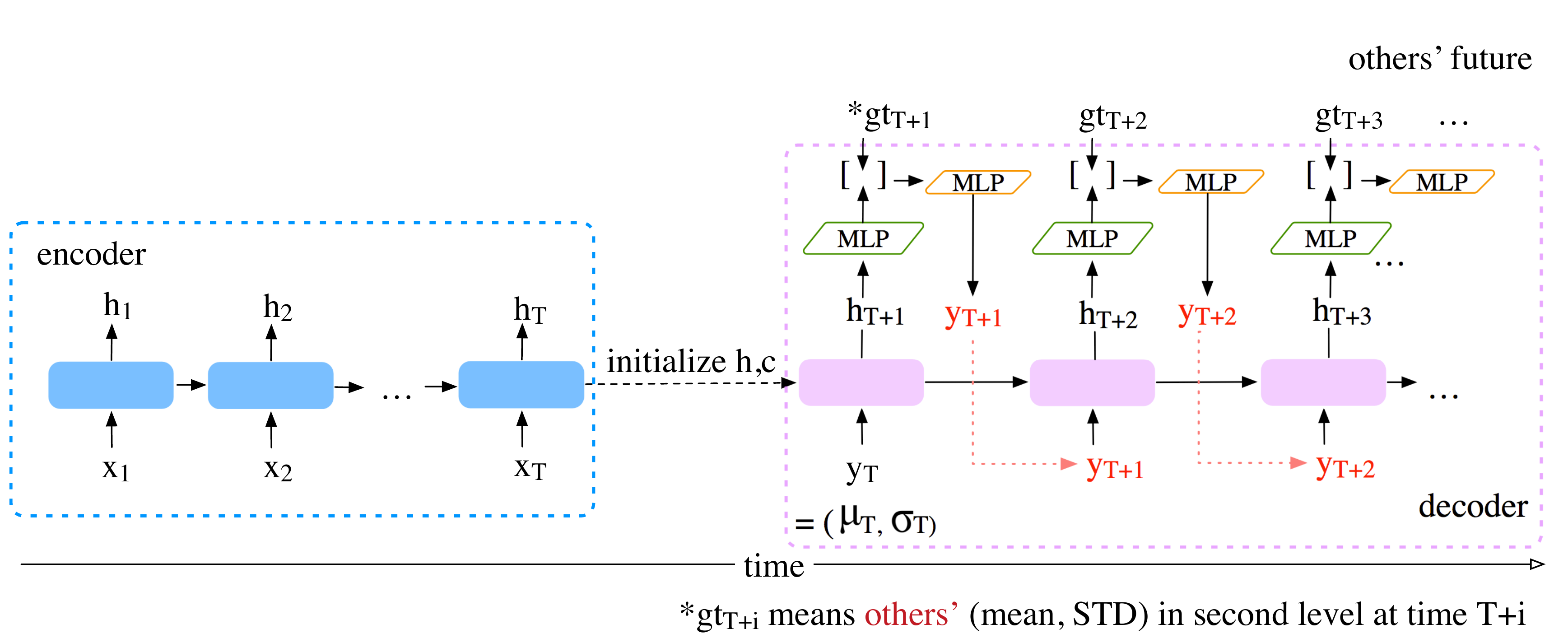}
   \caption{Mixing other users' information using a shared MLP mixing layer.}
\label{mlpmixing_fig}
\end{figure}

%------------------------------------------------------------------------- 
\textbf{MLP mixing.}
\label{mlpmixing} In this model (see Figure \ref{mlpmixing_fig}), we make use of other users' FoV at time $t$ when predicting the target user FoV at the same time.  We first use the sequence-to-sequence model in section \ref{seq2seq section} to predict the mean and STD of FoV centers, we then concatenate this temporary prediction with the FoV center means and STDs of other users at time $t$, and pass the concatenated predictions into the final projection layer (orange MLP in Figure \ref{mlpmixing_fig}) to get the final prediction. The last projection layer learns a mixing weight to combine target user's own prediction and others' known locations.

%------------------------------------------------------------------------- 

\textbf{Attentive Mixture of Experts (AME).} Inspired by works in the mixture of experts\cite{ame}, we can treat the trajectories of other users as experts' advice. When predicting for the target user, the known trajectories from other users can serve as guidance. We propose a novel attentive mixture of experts (AME)\cite{ame} module on the decoder part of the sequence to sequence model.  A context vector $c_i$ for each expert will be generated from the observations available for that expert $i$, which are then used to form a total context vector $c_{total} = \sum_i  \alpha_i c_i $. The weights $\alpha_i$ depend on the similarity of the target user and each expert. From $c_{total}$, we can then predict a property of the target user.

We consider two ways to derive the context vectors. In the first case, we directly use the FoV center locations (mean and STD of FoVs) of other users, denoted by $x_i$, as the context vectors. A shared embedding layer is applied to each $x_i$ and $x_{tar}$ to generate embedded feature $u_i$ and $u_{tar}$, where $x_{tar}$ is the temporary prediction generated by the seq2seq model. Finally, the prediction for the target user is generated by a weighted sum of $x_i$ using the similarity weighting, determined based on the embedded features $u_i$ and $u_{tar}$. We treat the initial prediction $x_{tar}$ as one of the experts so that the final predicted location is based on the target user's past FoVs as well as other users current FoVs. In the second case, we model each other user by a shared LSTM and treat the hidden states as the context vectors. A shared embedding layer is used to embed the hidden state $h_i$ to $u_i$ for each other user (at time $t$). We also use the same embedding layer to embed target user's hidden state  $h_{tar}$ to  $u_{tar}$. Similar to the first case, the final prediction is the weighted sum of the location $x_i$'s using the similarity score between $u_i$ and $u_{tar}$.

The similarity function $S(u_{tar},u_j)$ can take different forms as described in \cite{luong2015effective}, such as dot product, concatenation or MLP. We used the dot product in our experiments for simplicity. Note that using the hidden state to define the similarity can be interpreted as measuring the similarity between a user $i$ and the target user's trajectories \textit{so far}.

 \begin{figure}[t]
 \centering
   \includegraphics[width=1.0\linewidth]{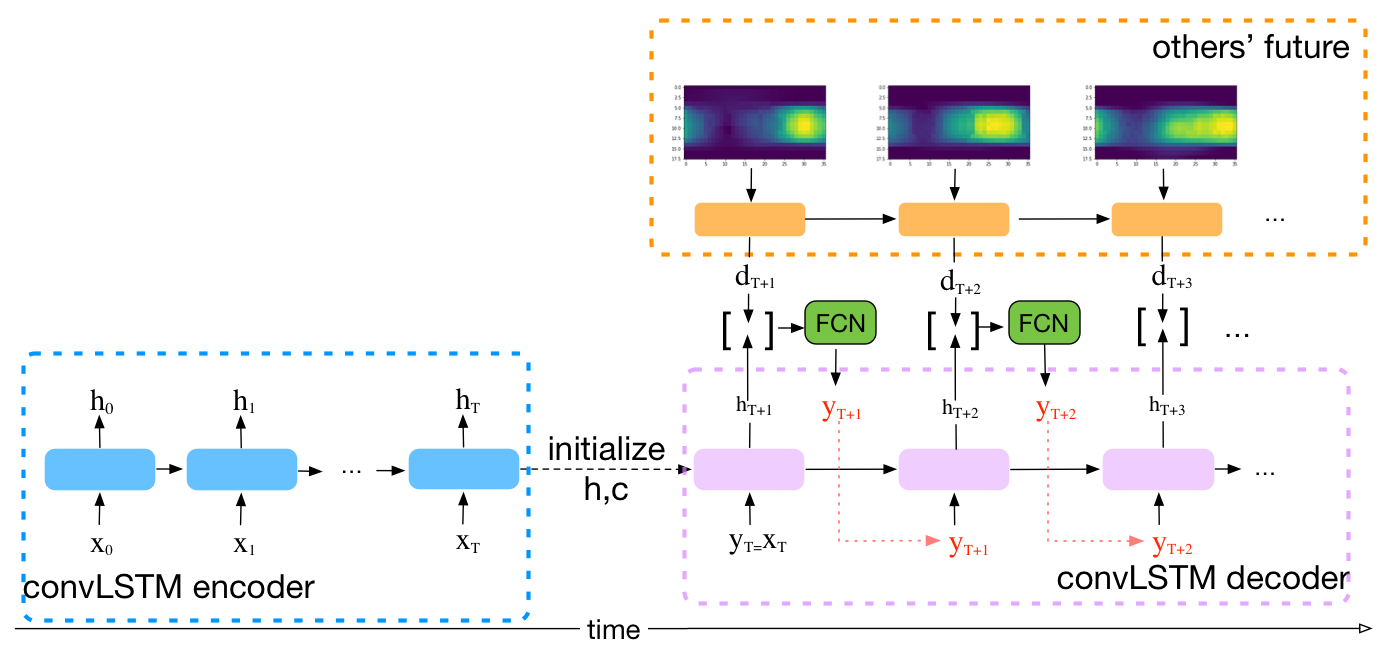}
   \caption{The model structure for predicting the future heatmap sequence for a target user using the target user's past heatmap sequence as well as the average heatmap sequence of other users. Each heatmap sequence is modeled using a convLSTM. All heatmaps are at second level.}
\label{convlstm_oth}
\end{figure}

%------------------------------------------------------------------------ 

\section{Prediction using 2D Equirectangular Heatmap Representation}

A potential problem of the trajectory-based models is that the mean and STD of FoV center within each second does not fully describe the FoV center distribution. It also does not explicitly exploit the fact that the FoV center is located on a spherical surface (a 2D plane in the equirectangular coordinate). To circumvent these problems, we further explore the use of a 2D heatmap representation of the FoV distributions within each second and model the dynamics of the heatmap sequences. 

Similarly, with the different variants of the trajectory-based model, we experimented with two kinds of models: predicting the future heatmap sequence from the target user's own past heatmap sequence and predicting using both the target user's past heatmap sequence as well as other users' heatmap sequences. We further considered using the saliency maps derived from the video content to help the prediction.

\subsection{Heatmap representation}

\begin{figure}[t]
 \centering
   \includegraphics[width=1\linewidth]{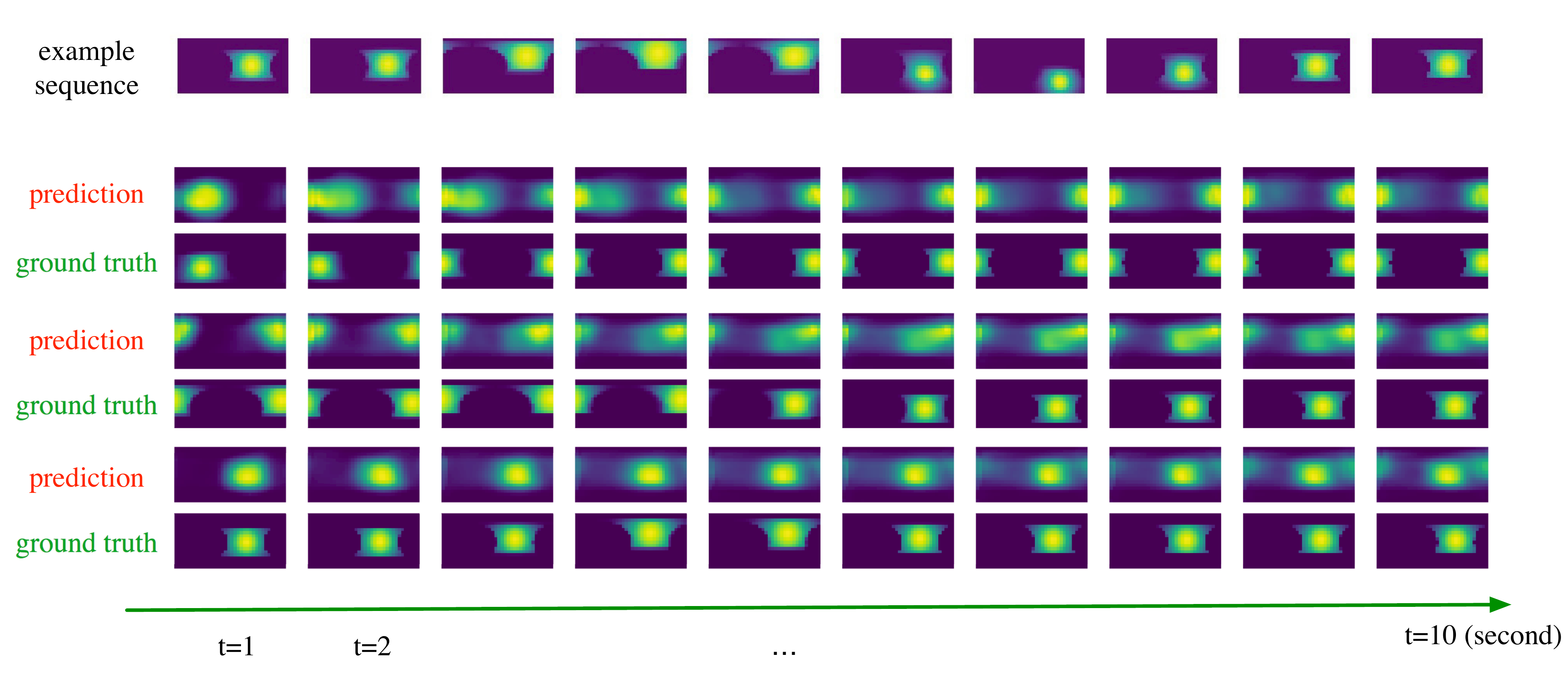}
   \caption{Heatmap-based method: example prediction results for the future 10 seconds.}
\label{2d_result}  
\end{figure}

Given the FoV center $(\theta, \phi)$ for a frame, we generate a 2D Gaussian heatmap in the equirectangular coordinate for this frame, where the horizontal axis represents longitude angle $\theta$ ranging from $-\pi$ to $\pi$, and the vertical axis represents latitude angle $\phi$ ranging from  $-\frac{\pi}{2}$ to $\frac{\pi}{2}$. To prevent the size of the heatmap from being too large for training, we discretize the $\theta-\phi$ plane by setting the bin size to $10\times 10$ degrees. Thus the size of the Gaussian heatmap is $18\times 36$. We assume the FoV spans $120^\circ \times 90^\circ$ so that each FoV initially corresponds to a $12\times9$ rectangle with value 1 in the heatmap. Row $i$ is then blurred by applying a Gaussian window with standard deviation $\sigma \propto \frac{1}{cos(\phi_i)}$, to account for the equirectangular projection distortion. Finally, we sum up all 30 frame-level heatmaps within one second to get the second-level heatmap (FPS=30). With this representation, an FoV center trajectory is described by a heatmap sequence. Example ground truth heatmap sequences are shown in Figure \ref{2d_result}.

\subsection{Prediction based on users' own heatmaps}
For using the target user information only, we used a seq2seq model where the encoder and decoder each uses a convolutional LSTM (convLSTM) (see the lower branch of Figure \ref{convlstm_oth}).  Both encoder and decoder contain 3 layers, generating 128, 64, and 32 channel hidden-state feature maps respectively, each with the same spatial size as the input heatmaps. A fully convolutional network (FCN) is then applied to the concatenated hidden state maps from all three layers, to generate a predicted heatmap for one second.

 \begin{figure}[t]
 \centering
\includegraphics[width=1.0\linewidth]{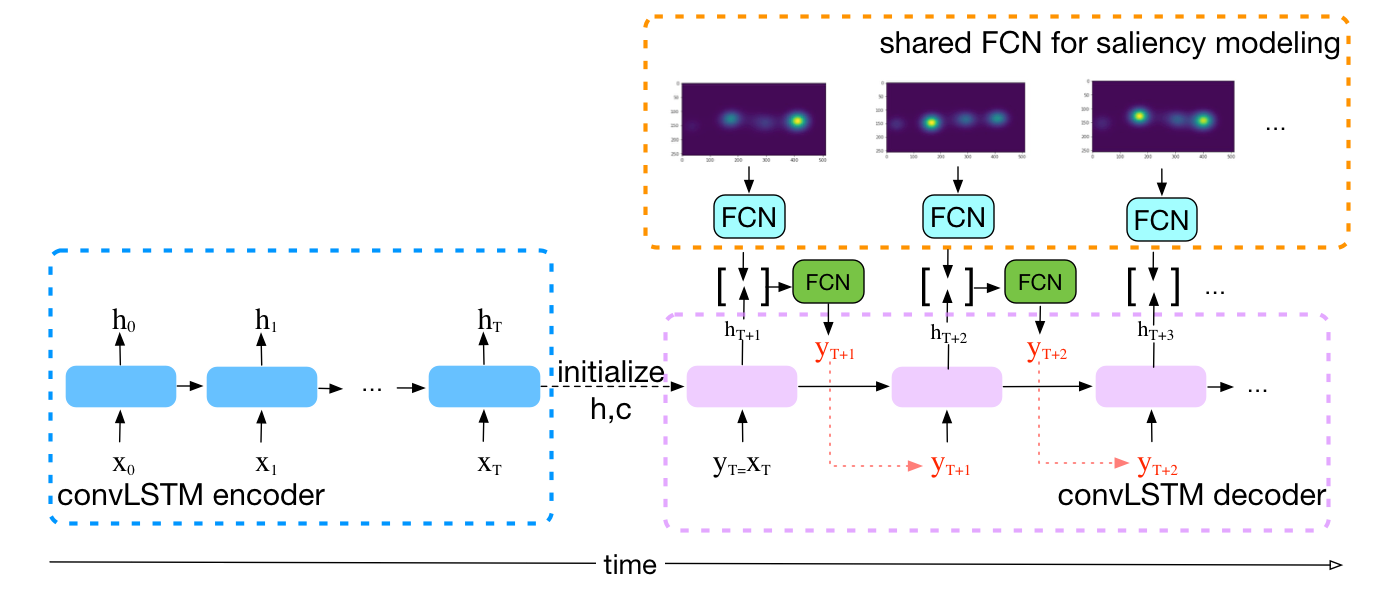}
\caption{Saliency heatmaps are modeled by a shared FCN at each time step. Saliency features are fused with the hidden states of the decoder convLSTM. }
\label{convLSTM_sal}
\end{figure}

\begin{figure*}[!ht]
\subfigure{\includegraphics[width=0.45\textwidth]{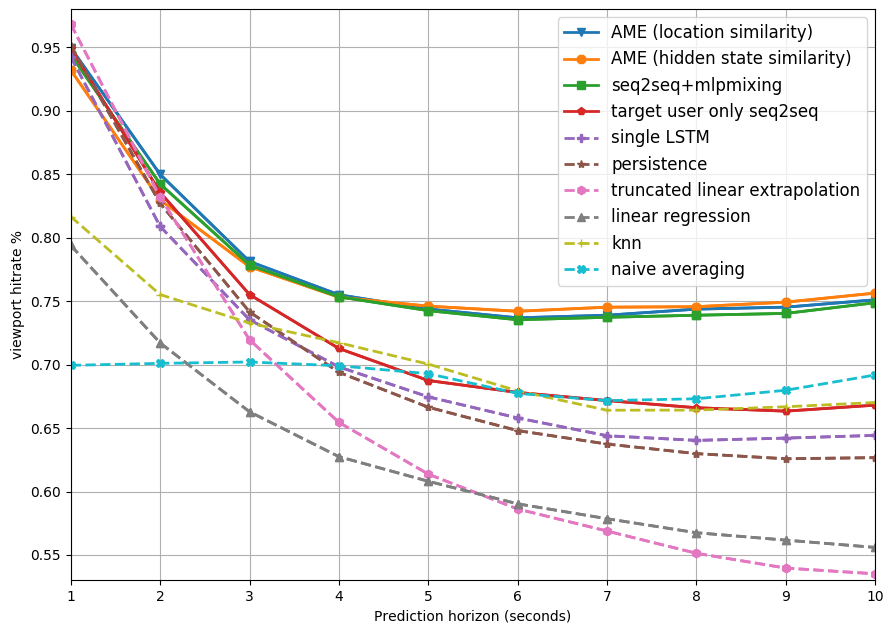}}
\hspace{0.05\textwidth}
\subfigure{\includegraphics[width=0.45\textwidth]{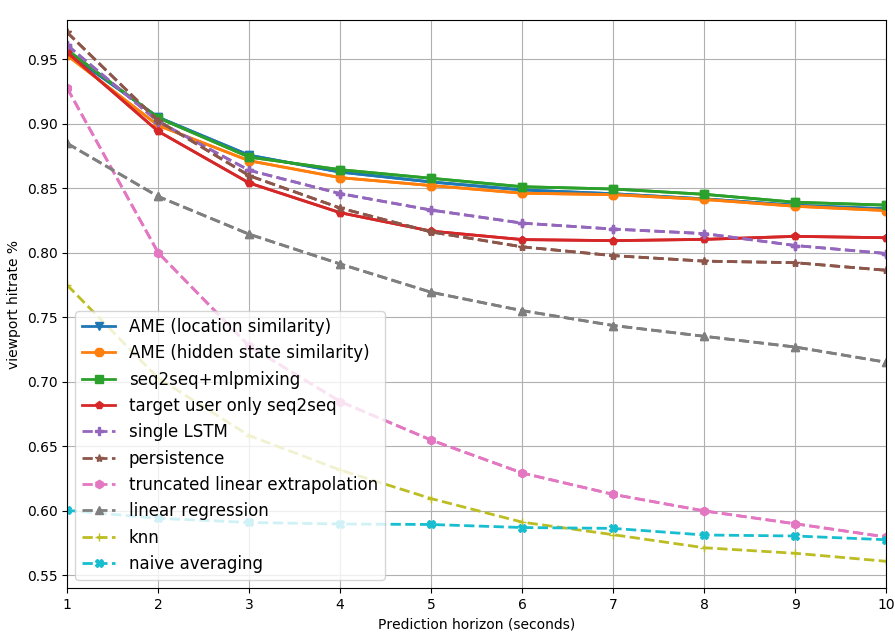}}
\caption{Hit rate of future 10 seconds for dataset \cite{shanghaitech} (left) and \cite{wu2017dataset} testset (right) for trajectory-based models, with spanning factor $\alpha=1.25$.}
\label{trj_curves}
\end{figure*}

\subsection{Utilizing other users' heatmaps} 
We explored different approaches to fuse others' information. 1) We model others' average heatmap sequence\footnote{Note that if the number of other users is large enough, the average heatmaps from other users at each second can be seen as a surrogate for the saliency map for this second. } using one convLSTM. At time $t+1$, the hidden states $d_{t+1}$ from this convLSTM and the hidden states $h_{t+1}$ from target user's decoder are concatenated and used to predict target user's heatmap at time $t+1$, $y_{t+1}$, through an FCN as shown in Figure \ref{convlstm_oth}. 2) Similar with the mlp mixing model in section \ref{mlpmixing}, we directly concatenate other users' heatmaps at time $t+1$ with the hidden states from target user's decoder and use the FCN to generate the final prediction. The FCN automatically learns the mixing weights (kernels) in order to generate the final predictions.

\subsection{Utilizing saliency maps derived from video sequences}
We applied a 2D saliency model\cite{sam} pretrained on regular 2D natural images, on the optical flow motion field images directly derived from the equirectangular images.  A total of 30 saliency maps are generated for each second, one for each frame. The saliency maps at time $t+1$ are fed into an FCN and the FCN features are then fused with the hidden states from the target user's convLSTM decoder. The combined features are then used to generate the predicted heatmap $y_{t+1}$, through another FCN, as shown in Figure \ref{convLSTM_sal}. We further explored using other users' FoV information and the visual saliency information together with the target user's past FoV information. Specifically, the hidden states of the target user decoder, the hidden states of the other users' convLSTM, as well as the visual saliency FCN feature maps are all concatenated to predict the future heatmap for the target user.

\section{Experiments}
%------------------------------------------------------------------------- 
We evaluate our models on two public datasets \cite{shanghaitech} and \cite{wu2017dataset}. We split each dataset into train and test subsets, each contains different videos. We train all our models using the train subset of \cite{shanghaitech} as it contains more dynamic scenes, and evaluate the models on both test subsets. Directly training and testing on dataset\cite{wu2017dataset} would have even higher performance, hence are omitted here. For the trajectory-based models, we use the Cartesian coordinate $(x,y,z)$ of the FOV center to characterize the FOV location. Note that we choose not to use the longitude $\theta$ and latitude $\phi$ angles to avoid the issue of $2\pi$ periodicity of $\theta$.

\subsection{Evaluation metrics}

\textbf{Hit rate.} The proposed trajectory-based FoV prediction methods are intended to be integrated into a viewport-based streaming system such as \cite{fanyi_MMSys2018}, where video segments in each second are precoded into different viewports covering different portions of the sphere surface. Based on the mean and STD of the predicted FoV centers for a future second, the server will send a viewport centered at the predicted mean (converted from $(x,y,z)$ to $(\theta, \phi)$). Generally, the viewport should cover a larger angle span than the FoV for every frame to accommodate the likely FoV shift within a second. Ideally, the angle span should be proportional to the predicted FoV standard deviation. However, this would require the server to precode and store viewports with different angle spans.  Here we consider a simpler system where the viewport's coverage area is fixed and is $\alpha^2$ times the area of the FoV of the HMD. For the trajectory-based model, we assume the FoV span is  $(120^\circ, 120^\circ)$ and we consider two expansion factors: $\alpha=1$ and $\alpha=1.25$, corresponding to the viewport angle span of $(120^\circ, 120^\circ)$ and $(150^\circ, 150^\circ)$ respectively. The hit rate of a viewport for each predicted second is the average percentage of the viewport's coverage area that is inside the actual frame FoV, for all the frames in that second.

%------------------------------------------------------------------------- 
\textbf{Mean Squared Error.} We also report the mean squared error between the predicted FoV center mean position in $(x,y,z)$ and the ground truth mean, averaged over the prediction horizons from 1 to 10 seconds.  

\textbf{Tile overlapping ratio}. Recall that the heatmap-based approach is intended for tile-based streaming systems, where the clients can request multiple tiles based on the predicted FoV heatmap. Therefore, we use the tile overlapping ratio as the performance metric for evaluating heatmap-based approaches. First, we determine the total number of bins $\text{Nbin}_\text{gt}$ in ground truth FoV heatmap that has non-zero values within that second, (recall that each pixel in the heatmap represents a bin with angle span $10^\circ \times10^\circ$). Next, we sort the confidence scores of each bin in the predicted heatmap and determine overlapping bins between the $\text{Nbin}_\text{gt}$ largest bins in the predicted heatmap and the ground truth heatmap. The ratio of the number of overlapping bins and $\text{Nbin}_\text{gt}$ is the tile overlapping ratio.

\begin{figure*}[!ht]
\subfigure{\includegraphics[width=0.45\textwidth]{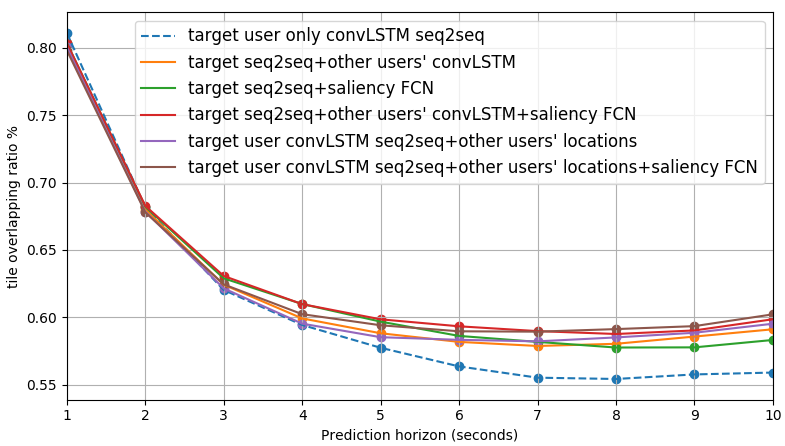}}
\hspace{0.05\textwidth}
\subfigure{\includegraphics[width=0.45\textwidth]{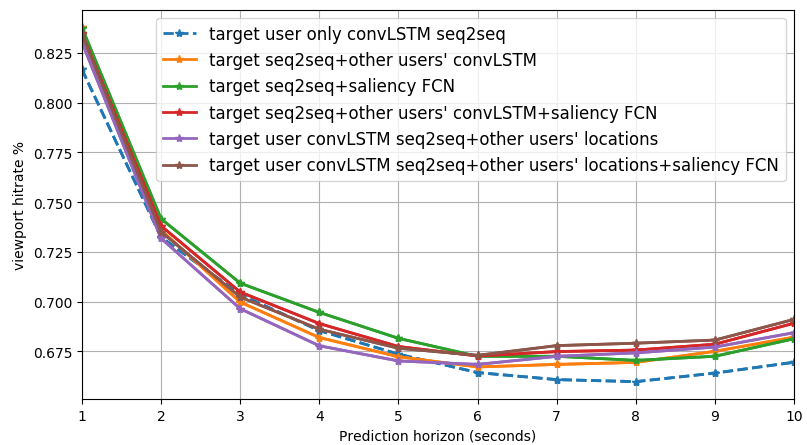}}
\caption{Heatmap-based models: Tile overlapping ratio (left) and hit rate (right) of future 10 seconds for dataset \cite{shanghaitech} testset.}
\label{2d_curve}
\end{figure*}

\textbf{FoV center estimation from the predicted heatmap}. To enable comparison between the heatmap-based approaches and the trajectory-based approaches, we also determine the mean location of the FoV centers from the predicted heatmap for each second. Based on this estimated location, we compute the hit rate of the corresponding viewport and also the MSE.  We determine the mean location by treating the normalized heatmap value in each pixel ($\theta, \phi$) as the probability that the FoV center is located at ($\theta, \phi$). The estimated mean location is then the weighted sum of all locations using the probability values as weights. What's more, recognizing that the number of effective pixels along the line at the latitude $\phi$ decreases with a factor of $\cos(\phi)$ ($\phi\in [-\frac{\pi}{2},\frac{\pi}{2}]$), we weight the contribution of the pixels at ($\theta,\phi$) by $\cos(\phi)$. We also take care of the $2\pi$ periodicity of the longitude $\theta$ when computing the mean.

\begin{table}[!ht]
\fontsize{6}{7.5}\selectfont
\begin{center}
\centering
\caption{Performances of various Trajectory-based models.}  % (after scaling $\theta$)
\label{table_trj}
\begin{tabular}{c| c|c| c|c|c| c}
\hline
&\multicolumn{3}{ c| }{Shanghai Dataset\cite{shanghaitech}} &\multicolumn{3}{ c }{Tsinghua Dataset\cite{wu2017dataset}} \\
\hline
&\multicolumn{2}{ c| }{Average Hit Rate} & MSE& \multicolumn{2}{ c| }{Average Hit Rate} & MSE \\
\textbf{Model Variants}& a=1.25&a=1 & & a=1.25&a=1  &\\
\hline 
linear regression&0.6260 & 0.4995 &1.3831 &0.7777 &    0.6572   &0.7479 \\
\hline
truncated linear &&&&&&\\
extrapolation& 0.6565  &  0.5297   &1.5197 &0.6802 & 0.5424 & 1.5368\\
\hline
{persistency}& 0.7043  & 0.5922 &0.9210 &  0.8355 &0.7398   &0.4691\\
\hline
{KNN (k=5)}& 0.7025 & 0.5775 &  0.8982 & 0.6199 & 0.5171 &0.8635\\
\hline
{Naive Average}& 0.6880 & 0.5586 & 1.1377 & 0.5873 &  0.4878 &0.7717\\
\hline
{single LSTM}& 0.7083   & 0.5834  &0.6164 &0.8464  &0.7346  &0.3731 \\
\hline
{target user only seq2seq}&  0.7283  &0.6049   &0.5881 & 0.8402  &0.7253  &0.3853\\
\hline
{seq2seq+mlpmixing} &  0.7757  & 0.6510  &0.4890 &\textbf{0.8677 }& 0.7518  & 0.3043\\

\hline
AME &&&&&&\\
(location similarity)&\textbf{ 0.7791 }& \textbf{0.6566 }& \textbf{0.4807} & 0.8658 &\textbf{ 0.7507}&\textbf{0.3037}\\
\hline
 AME &&&&&&\\
 (hidden state similarity) &0.7772 & 0.6552   & 0.4983 &0.8632 & 0.7474 & 0.3273 \\
\hline
\end{tabular}
\end{center}
\end{table}

\subsection{Perfomance comparison}
For the trajectory-based methods, we compare our proposed models with several baseline methods: \textbf{1}. persistency (repeating the location of the FoV center in the last frame in the past), \textbf{2}. linear regression on the last 10 seconds to predict, \textbf{3}. truncated linear extrapolation (linear regression on the last monotonic line segment), \textbf{4}. Naive average: averaging all other user's locations at time $t$ as the prediction for time $t$. \textbf{5}. K nearest neighbors (KNN): selecting K out of all other users at time $t$ who are closest to the target user's predicted position at time $t-1$,  and use the average of these K positions as the prediction for time $t$. We used K=5. \textbf{6}. single LSTM, which uses the same LSTM model for the past and future. These baselines are common choices in related works (section \ref{relatedworks}).

Figure \ref{trj_curves} show the hit rate curves for trajectory-based models. We can see that the hit rate of the persistency and truncated linear extrapolation models drop very rapidly as the prediction horizon increases, indicating the nonlinear nature of FoV trajectories. Our proposed model using the target user information only (target user only seq2seq) outperforms all baselines by a large margin. Furthermore, our models that utilize other users' information yield much higher hit rates at prediction horizons between 4-10 seconds. However, different ways of utilizing other users' information lead to very similar performance; Modeling others using an LSTM and using the hidden state similarity between others and the target user (AME (hidden state similarity)) does not provide gains over just using other users location information at the prediction time (AME (location similarity) and seq2seq+mlpmixing). But compared with KNN and Naive average, our model learns automatically different mixing weights for other users based on their similarities with the target user, leading to much better performances. Table \ref{table_trj} compares the trajectory-based methods in terms of the average hit rate and the MSE. Overall, the two methods of using others' trajectory information based on location similarity (AME (location similarity) and seq2seq+mlpmixing) have the best performance.

In Figure \ref{2d_curve} and Table \ref{table_2d}, we compare the performances of heatmap-based models. Overall, exploiting other users' heatmaps and the saliency feature maps give the best performance. However, different ways of utilizing others' information yield very similar performances. Furthermore, the gap between utilizing both others' heatmaps and saliency maps v.s. utilizing only one of these is rather small. This suggests that the saliency information and the information from others' average heatmaps are not orthogonal. 
 
Comparing the hit rate and MSE achievable by the heatmap-based methods (Table \ref{table_2d}) with those obtained by the trajectory-based models (Table \ref{table_trj}),  we see that the trajectory-based approaches are significantly better for predicting the mean of the FoV centers in each second.  Such mean prediction is desirable for viewport-based streaming, where the system can only deliver a continuous viewport for each future second, and the center and span of the viewport needs to be determined, to maximally cover all the FOVs over the entire second. For tile-based streaming  systems, the heatmap-based approaches may be more appropriate, as it predicts the FoV center distribution. For example, when a predicted heatmap includes multiple separate peaks, the system can send multiple non-contiguous tiles, corresponding to different peaks.

\section{Conclusion}
In this paper, we proposed two groups of FoV prediction models: trajectory-based models and heatmap-based models to suit different needs in viewport-based streaming and tile-based streaming scenarios respectively. For each group, we further considered models 1) using the target user's information only, or 2) utilizing other users' FoVs as well. For heatmap-based models, we also considered utilizing the visual saliency information. We proposed multiple model variants in both groups, especially, the MLP mixing model and the Attentive Mixture of Experts (AME) model in the trajectory-based group to automatically learn the importance weights of other users' contributions to the final prediction. For heatmap-based models, we explored several ways to fuse the features from the users' FoV heatmaps as well as from the video content. We have evaluated the proposed models on two public datasets and showed that the proposed models utilizing the target user's past information have higher accuracies in long term predictions (4-10 seconds ahead) than popular baseline methods in the literature. Furthermore, models utilizing other users' information provide substantial performance gain over utilizing the target user's information only.

\begin{table}[!t]
\fontsize{6.5}{8}\selectfont
\begin{center}
\centering
\caption{Performances of various heatmap-based models.}
\label{table_2d}
\begin{tabular}{c| c|c|c}
\hline
&\multicolumn{3}{ c }{Shanghai Dataset\cite{shanghaitech}}\\
\hline
& Average tile  &Average  &  \\
\textbf{Model Variants}&overlapping   &Hit Rate &MSE  \\
&ratio& (a=1.25) &\\
\hline
target user only convLSTM seq2seq &0.5987 & 0.6943  &  0.9477\\
\hline
seq2seq + others' convLSTM &0.6127  &0.7059   &  0.9100\\
\hline
seq2seq + saliency FCN &0.6148  &0.7079  &   0.8977\\
\hline
seq2seq + others convLSTM &&&\\
and saliency FCN & \textbf{0.6203} &0.7097   &  0.8939\\
\hline
seq2seq + mlpmixing others' heatmaps &0.6130 &\textbf{0.7136} &  \textbf{0.8797}\\
\hline
seq2seq + mlpmixing (others'  &&& \\
heatmaps and saliency fcn features) &0.6180  &0.7101     &  0.8917\\
\hline
\end{tabular}
\end{center}
\end{table}

\bibliographystyle{IEEEtran}
\bibliography{fov_arxiv}

% Generated by IEEEtran.bst, version: 1.13 (2008/09/30)
\begin{thebibliography}{10}
\providecommand{\url}[1]{#1}
\csname url@samestyle\endcsname
\providecommand{\newblock}{\relax}
\providecommand{\bibinfo}[2]{#2}
\providecommand{\BIBentrySTDinterwordspacing}{\spaceskip=0pt\relax}
\providecommand{\BIBentryALTinterwordstretchfactor}{4}
\providecommand{\BIBentryALTinterwordspacing}{\spaceskip=\fontdimen2\font plus
\BIBentryALTinterwordstretchfactor\fontdimen3\font minus
  \fontdimen4\font\relax}
\providecommand{\BIBforeignlanguage}[2]{{%
\expandafter\ifx\csname l@#1\endcsname\relax
\typeout{** WARNING: IEEEtran.bst: No hyphenation pattern has been}%
\typeout{** loaded for the language `#1'. Using the pattern for}%
\typeout{** the default language instead.}%
\else
\language=\csname l@#1\endcsname
\fi
#2}}
\providecommand{\BIBdecl}{\relax}
\BIBdecl

\bibitem{xingjian2015convolutional}
S.~Xingjian \emph{et~al.}, ``Convolutional lstm network: A machine learning
  approach for precipitation nowcasting,'' in \emph{Advances in neural
  information processing systems}, 2015, pp. 802--810.

\bibitem{bao2016shooting}
Y.~Bao \emph{et~al.}, ``Shooting a moving target: Motion-prediction-based
  transmission for 360-degree videos.'' in \emph{BigData}, 2016, pp.
  1161--1170.

\bibitem{fan2017fixation}
C.-L. Fan \emph{et~al.}, ``Fixation prediction for 360 video streaming in
  head-mounted virtual reality,'' in \emph{Proceedings of the 27th Workshop on
  Network and Operating Systems Support for Digital Audio and Video}.\hskip 1em
  plus 0.5em minus 0.4em\relax ACM, 2017, pp. 67--72.

\bibitem{shanghaitech}
Y.~Xu \emph{et~al.}, ``Gaze prediction in dynamic 360 immersive videos,'' in
  \emph{Proceedings of the IEEE Conference on Computer Vision and Pattern
  Recognition}, 2018, pp. 5333--5342.

\bibitem{xu2018predicting}
M.~Xu \emph{et~al.}, ``Predicting head movement in panoramic video: A deep
  reinforcement learning approach,'' \emph{IEEE transactions on pattern
  analysis and machine intelligence}, 2018.

\bibitem{ban2018cub360}
Y.~Ban \emph{et~al.}, ``Cub360: Exploiting cross-users behaviors for viewport
  prediction in 360 video adaptive streaming,'' in \emph{2018 IEEE
  International Conference on Multimedia and Expo (ICME)}.\hskip 1em plus 0.5em
  minus 0.4em\relax IEEE, 2018, pp. 1--6.

\bibitem{tilebased}
Z.~Xu \emph{et~al.}, ``Tile-based qoe-driven http/2 streaming system for 360
  video,'' \emph{IEEE ICME Grand Challenge on DASH}, 2018.

\bibitem{xie2018cls}
L.~Xie \emph{et~al.}, ``Cls: A cross-user learning based system for improving
  qoe in 360-degree video adaptive streaming,'' in \emph{2018 ACM Multimedia
  Conference on Multimedia Conference}.\hskip 1em plus 0.5em minus 0.4em\relax
  ACM, 2018, pp. 564--572.

\bibitem{sutskever2014sequence}
I.~Sutskever \emph{et~al.}, ``Sequence to sequence learning with neural
  networks,'' in \emph{Advances in neural information processing systems},
  2014, pp. 3104--3112.

\bibitem{ame}
P.~Schwab \emph{et~al.}, ``Granger-causal attentive mixtures of experts:
  Learning important features with neural networks. arxiv preprint,''
  \emph{arXiv preprint arXiv:1802.02195}, 2018.

\bibitem{luong2015effective}
M.-T. Luong \emph{et~al.}, ``Effective approaches to attention-based neural
  machine translation,'' \emph{arXiv preprint arXiv:1508.04025}, 2015.

\bibitem{wu2017dataset}
C.~Wu \emph{et~al.}, ``A dataset for exploring user behaviors in vr spherical
  video streaming,'' in \emph{Proceedings of the 8th ACM on Multimedia Systems
  Conference}.\hskip 1em plus 0.5em minus 0.4em\relax ACM, 2017, pp. 193--198.

\bibitem{sam}
M.~Cornia \emph{et~al.}, ``Predicting human eye fixations via an lstm-based
  saliency attentive model,'' \emph{arXiv preprint arXiv:1611.09571}, 2016.

\bibitem{fanyi_MMSys2018}
\BIBentryALTinterwordspacing
L.~Sun \emph{et~al.}, ``Multi-path multi-tier 360-degree video streaming in 5g
  networks,'' in \emph{Proceedings of the 9th ACM Multimedia Systems
  Conference}, ser. MMSys '18.\hskip 1em plus 0.5em minus 0.4em\relax New York,
  NY, USA: ACM, 2018, pp. 162--173. [Online]. Available:
  \url{http://doi.acm.org/10.1145/3204949.3204978}
\BIBentrySTDinterwordspacing

\end{thebibliography}
\end{document}